\documentclass{article}

\usepackage{arxiv}

\usepackage[utf8]{inputenc} 
\usepackage[T1]{fontenc}    
\usepackage{hyperref}       
\usepackage{url}            
\usepackage{booktabs}       
\usepackage{amsmath,amsfonts}
\usepackage{nicefrac}       
\usepackage{microtype}      
\usepackage{graphicx}
\usepackage{subfig}
\usepackage{graphicx}
\usepackage{natbib}
\usepackage{doi}
\usepackage{acro}
\usepackage{multirow}
\usepackage{algorithm}
\usepackage{algpseudocode}

\title{Camera clustering for scalable stream-based active distillation}

\author{ \href{https://orcid.org/0000-0001-9034-0794}{\includegraphics[scale=0.06]{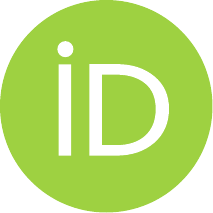}\hspace{1mm}Dani Manjah}\thanks{Dani Manjah and Benoit Macq are affiliated with
Université catholique de Louvain, 
Belgium. Christophe De Vleeschouwer is funded by the  National Fund for Scientific Research (FNRS). Davide Cacciarelli is affiliated with Imperial College London. 
\textbf{16/04/2024 - This manuscript is currently under review at IEEE Transactions on Circuits and Systems for Video Technology.}} \\
	\texttt{dani.manjah@uclouvain.be} \\
	\And
	\href{https://orcid.org/0000-0001-6664-9038}{\includegraphics[scale=0.06]{orcid.pdf}\hspace{1mm}Davide Cacciarelli} \\
 \texttt{d.cacciarelli@imperial.ac.uk} \\
 	\And
	\href{https://orcid.org/0000-0001-5049-2929}{\includegraphics[scale=0.06]{orcid.pdf}\hspace{1mm}Christophe De Vleeschouwer} \\
  \texttt{christophe.devleeschouwer@uclouvain.be} \\
  	\And
	\href{https://orcid.org/0000-0002-7243-4778}{\includegraphics[scale=0.06]{orcid.pdf}\hspace{1mm}Benoît Macq} \\
 \texttt{benoit.macq@uclouvain.be} \\
}
\date{}
\DeclareAcronym{DNN}{
  short={DNN},
  long={Deep Neural Network}
}

\DeclareAcronym{CCTV}{
  short={CCTV},
  long={Closed-Circuit Television}
}

\DeclareAcronym{KD}{
  short={KD},
  long={Knowledge Distillation}
}

\DeclareAcronym{AL}{
  short={AL},
  long={Active Learning}
}

\DeclareAcronym{SBAD}{
  short={SBAD},
  long={Stream-Based Active Distillation}
}

\DeclareAcronym{UDA}{
  short={UDA},
  long={Unsupervised Domain Adaptation}
}

\DeclareAcronym{MC}{
  short={MC},
  long={Model Compression}
}
\hypersetup{
pdftitle={Camera clustering for scalable stream-based active distillation},
pdfsubject={GroupDistillation},
pdfauthor={Dani Manjah, },
pdfkeywords={CSBAD,Object Detection, Knowledge Distillation, Active Learning, Neural Network Deployment},
}
\begin{document}



\maketitle
\begin{abstract}
We present a scalable framework designed to craft efficient lightweight models for video object detection utilizing self-training and knowledge distillation techniques. We scrutinize methodologies for the ideal selection of training images from video streams and the efficacy of model sharing across numerous cameras. By advocating for a camera clustering methodology, we aim to diminish the requisite number of models for training while augmenting the distillation dataset. The findings affirm that proper camera clustering notably amplifies the accuracy of distilled models, eclipsing the methodologies that employ distinct models for each camera or a universal model trained on the aggregate camera data.
\end{abstract}
\keywords{Object Detection, Knowledge Distillation, Active Learning, Neural Network Deployment.}

\section{Introduction}
\label{sec:Introduction}
Traditional training methodologies for Deep Neural Networks (DNNs) face substantial barriers, including significant time, financial investment, and labor for dataset preparation and training \cite{LucasBeyer2022}. These challenges create limitations in operating and scaling ever-evolving video-analytics systems such as city-scale CCTV because 1) DNNs need to be regularly fine-tuned to maintain performance and 2) update costs increase with the number of cameras.  

To address these issues, we propose a large general-purpose model fine-tuning compact DNNs on subsets of cameras. Our method facilitates localized consistent updates, which are crucial for maintaining local DNN performance in individual cameras \cite{sbad} and promoting cost-efficient training and scalability \cite{whatisscal,recentscalabi,clusterscalabledef}. 
This paper builds on the Stream-Based Active Distillation (SBAD) mechanism, initially presented in \cite{sbad}, which selects images from a single stream on-the-fly for training a low-complexity DNN optimized for local conditions. We propose a novel multi-stream SBAD approach based on clustering camera nodes in a multi-camera environment. This method rests on the premise that models fine-tuned on a specific camera domain are likely to transfer effectively to other domains with similar characteristics.

\begin{figure}[htpb]
    \centering
    \includegraphics[width=0.5\linewidth]{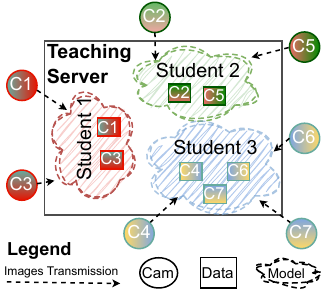}
    \caption{Updates of video analytics models: Cameras (C1 to C7) send selected image data to a central server, which pseudo-labels, trains and updates specialized \emph{Students} models for groups of similar cameras. These updated models are then sent back to their associated cameras.} 
    \label{fig:introClustering}
\end{figure}

Our cluster-based fine-tuning (Fig. \ref{fig:introClustering}) reduces the marginal cost of training additional units and enhances DNN accuracy by striking a balance between specificity (allocating a model for each camera) and universality (utilizing a larger, diverse training set). In other words, stream aggregation not only reduces the requirement for numerous student models but also improves their prediction accuracy compared with training a separate model for each stream with only its specific images. Moreover, this method achieves higher accuracy than training a universal student model with images from all streams, indicating that employing multiple clusters leads to superior accuracy.
\newpage
Our key contributions include:
\begin{enumerate}
\item The validation of active distillation in a multi-camera setup. We showed that selecting the frames obtaining the highest confidence score by local models results in the most accurate DNN. This includes an analysis of the potential biases of model-based pseudo-annotations.
\item Introduction of a novel camera-clustering based on model cross-performance, coupled with an in-depth analysis of its impact on model accuracy. This dual contribution showcases model compactness improvements without sacrificing accuracy and explores how various clustering strategies affect training complexity and model performance.
\item Novel data and codebase available  at \url{https://github.com/manjahdani/CSBAD/} to support further research.
\end{enumerate}

\section{Related Work}
\label{sec:sota}

We previously introduced Stream-Based Active Distillation (SBAD), a technique combining Knowledge Distillation (KD) with Active Learning (AL) to train compact DNN models tailored for their video-stream \cite{sbad}. We had developed sampling techniques for unlabeled streaming data, mitigating labeling inaccuracies and aiming to assemble the smallest training set that achieves the best local performance. This paper expands upon this by proposing Clustered Stream-Based Active Distillation (CSBAD), which organizes cameras' data into clusters; balancing training complexity and improving model accuracy. This section delves into foundational principles of SBAD — specifically, KD and AL — and reviews the existing literature on scene clustering.



\subsection{Knowledge Distillation}
\acs{KD} is a process where a compact \emph{Student} model learns from a more complex \emph{Teacher} model \cite{hinton2015distilling}. 
It is used to 
create lightweight models that are suited to limited storage and computational capacities \cite{cheng2017survey}, as encountered on edge devices \cite{mishra2023designing,tanghatari2023federated}. Through regular updates, it can ensures that the \emph{Student} model continues to perform effectively even when operating on changing data distributions \cite{saito2019semi}.


Recently, \acs{KD} has supported innovative applications in video analytics, particularly through \textit{Online Distillation} \cite{cioppa2019arthus,mullapudi2019online}, where a compact model's weights are updated in real time to mimic the output of a larger, pre-existing model. Other notable applications include \textit{Context- and Group-Aware Distillation} \cite{rivas2022towards,delta_distillation_2022,vilouras2023group}, which accounts for context and/or sub-population shifts when delivering student models. 
Finally, \textit{Active Distillation} \cite{baykal2022robust} combines AL with KD to train students with a reduced number of appropriately selected samples, while \textit{Robust Distillation} \cite{goldblum2020adversarially} explores the transfer of adversarial robustness from \emph{Teacher} to \emph{Student}.


In our work, we studied the impact of using \acs{KD} instead of a human annotator on \emph{Students}' accuracy.  




\subsection{Active Learning} \label{sec:sota_al}
\acs{AL}
aims to select and label the most informative data points to reduce the number of samples required to train a model \cite{Settles2009}. \acs{AL} can be implemented offline (Pool-Based \acs{AL}) or on-the-fly (Stream-Based \acs{AL}), from continuous streams.

Stream-Based \acs{AL} is the most relevant for video-based systems, where vast amounts of unlabeled data arrive continuously, making data storage impractical. However, most of the \acs{AL} methods developed for \acp{DNN} have been focused on the pool-based scenario \cite{Sener2017,Yoo2019,Ash2019,Elhamifar2013,Agarwal2020,Prabhu,sinha2019variational,Yuan2021}. 
Stream-based \acs{AL} has mostly been studied in conjunction with classification or regression models \cite{SBAL,ROAL,ICML_SBAL}, with traditional statistical models \cite{ROZANEC2022277, Narr2016}. Those methods do not generalize to \acp{DNN}. 
In contrast, our work considers \acs{AL} to train a DNN with images labeled by a \emph{Teacher} model (and not a human). As an original contribution, we show that the confidence associated with automatic labeling of training samples plays a crucial role in the AL selection process.

\subsection{Scene Clustering}
In multi-camera systems, the technique of scene clustering plays a crucial role in aggregating and synthesizing information from diverse sources effectively. This technique supports video summarization \cite{aner2002video}, anomaly detection \cite{sun2023hierarchical}, and content recommendation \cite{wang2013intelligent}, with hierarchical methods addressing redundancy in tracking and overlapping scenes \cite{li2023hierarchical,szHucs2023multi,wang2008correspondence,specker2021occlusion,patel2021multi, simon2010visual}. Despite advances in techniques such as multiview clustering \cite{multiviewclustering,fastmultiviewclustering}, a significant challenge persists in \textbf{selecting representative frames for each video} and, to the best of our knowledge, no previous work has studied the aggregation of views to reduce the number and improve the accuracy of student models serving a multi-camera system. Our research introduces a novel scene-clustering approach that groups videos based on the cross-dataset performance of models with the aim of improving model training and system efficiency in multi-camera environments.

\section{Problem Statement}
\label{sec:method}


We focus on large-scale networks where nodes, equipped with compact DNNs, analyze video-streams such as city-scale CCTVs. All nodes are connected to a central training server. 
In our system, the central unit collects the images from the nodes, annotates them, and trains the DNNs to be deployed on the nodes. In practice, DNNs may require periodic updates or adjustments when underperforming due to changes in the operating conditions. Furthermore, the system can grow with the deployment of additional sources of video. Therefore, we seek a \textbf{cost-effective} \textbf{scalable} \cite{whatisscal,recentscalabi} training methodology for DNNs suited to the visual specificity of their video streams. 


Since automatic management of the system is desired \cite{clusterscalabledef}, annotation of the collected data is implemented using a large, general purpose \emph{Teacher} model instead of human annotators. 
In practice, the \emph{Teacher} may be inaccurate, potentially causing \emph{Students} DNNs to overfit to incorrect pseudo-labels. This phenomenon is generally referred to as \emph{confirmation bias} \cite{arazo2020pseudo}, and is further studied in Section \ref{sec:confirmation bias}.

Obviously, the amount of data collected per stream and the number of trained DNNs directly affect the maintenance (DNNs update) costs and system scalability. 
To mitigate training complexity in large-scale systems, we propose to cluster streams based on their similarities and to train a single model for each cluster.
Choosing the right number of clusters is not trivial. Indeed, fewer but larger clusters allow models to train on a broader diversity, enhancing adaptability \cite{chen2023ccsd}. Conversely, numerous but smaller clusters fine-tune models to the nuances of individual streams but are trained with less data \cite{sbad}. 

Given a fixed number of training images per stream, the partitioning in clusters also calls for reevaluation of the training management. Indeed, in DNN training, the number of times a model updates its weights in one epoch depends on the amount of training images. Hence, with a fixed number of training epochs, models trained over smaller clusters update their weights less frequently. This disadvantages those models compared with those trained on larger clusters. Therefore, in scenarios where computational power is abundant, we could investigate increasing the number of epochs to reach the model's peak accuracy even when the training set gets smaller. In this context, our work 
addresses four research questions:

\begin{enumerate}
    \item How can we select images from each stream to maximize the resulting model's accuracy?
    \item To what extent does the quality of pseudo-labels affect downstream models' performance? How can we mitigate the impact of inaccurate pseudo-labels and reduce the risk of confirmation bias? 
    \item Given a constant number of epochs per model, thereby keeping the overall training complexity constant, which partitioning results in the highest model performance?  
    \item Given a constant number of iterations (gradient descents) per model, which partitioning results in the highest model performance? 
\end{enumerate}




\section{Methodology}
\label{sec:methodology}

Consider a set \( \mathcal{S} \) of $N$ video streams \(\ \{\mathcal{X}_1,\cdots,\mathcal{X}_N\}\), with each \( \mathcal{X}_i \) linked to \textbf{a compact 
} \emph{Student} model, \( \theta_i \) for object detection. To maintain \emph{Students} up-to-date, models are fine-tuned on a central server using selected samples from the videos. The two-level system, encompassing \emph{Students} and a \emph{Teaching Server}, 
is detailed below. 
\subsection{Student-level}
\emph{Students} operate on cost-effective hardware and use a \texttt{SELECT} \( (I) \) function to decide whether an image \( I \) is informative to train the model. \texttt{SELECT} functions must be \emph{efficient}. Efficiency is crucial; if \texttt{SELECT} takes longer than the frame interval to make decisions, a buffer must be used to store incoming images. This increases the demand for data storage, thus conflicting with scalability goals. Additionally, to reduce training cost, the set of selected frames and their associated pseudo-labels, which constitutes the training set $\mathcal{L}_i$, must not exceed a frame budget $B$ per student, hence, $|\mathcal{L}_i|\leq B$, for all $i=1,\cdots,N$. 

\subsection{Teaching Server}
To moderate annotation costs, the images are pseudo-labeled by a universal but imperfect model $\Theta$ referred to as \emph{Teacher} \cite{originSelfDistillaion}
. Upon receiving an image $I$ from a stream $\mathcal{X}_i$, \( \Theta\) pseudo-labels it by generating a set of bounding boxes $\tilde{P}$ and then adds the pair \( (I,\tilde{P}) \) to the image stream's database $\mathcal{L}_i$, associated to the $i^{th}$ student. This process results in a collection of sets ${\mathcal{L}_1,\cdots,\mathcal{L}_N}$.
Next, the server employs a \texttt{CLUSTER} method that partitions the databases $\{\mathcal{L}_i\}_{i=1,...,N}$ into \( K \leq N \) training sets \( \{E_1, \ldots, E_K\} \). The \texttt{CLUSTER} approach builds on the premise that models are transferable between similar streams and is detailed in Section \ref{sec:CLUSTERmethod}. We also define a mapping function, \(map(\cdot)\), linking \emph{Students} to their clusters, thus enabling the appropriate updating of their weights post-training.


$K$ models are then trained on each $E_k$, with $k=1,\cdots,K$, and deployed on associated cameras. Algorithm \ref{alg:1} provides the 
pseudo-code of the method. 



\begin{algorithm}
\caption{CSBAD Framework}\label{alg:1}
\begin{algorithmic}[1]
\Require pre-trained student model $\theta$, a general purpose teacher model $\Theta$, a training frame budget $B$, a \texttt{SELECT} strategy, image streams $\{\mathcal{X}_1, \cdots, \mathcal{X}_N\}$, a mapping rule \texttt{CLUSTER}, K partitions.
\Ensure $B \geq 1$, $1 \leq K\leq N$.
\Statex
\Comment{\MakeUppercase{INITIALIZATION \& SAMPLING}}
\For{$\mathcal{X}_i \in \mathcal{S}$} \Comment{$i = 1,\cdots, N$}
    \State $\theta^{\mathcal{X}_i} \gets \theta^{general}$
    \State $\mathcal{L}_i \gets \emptyset $ 
    \State $t \gets 0$ \Comment{Timestamp}
    \While{$|\mathcal{L}_i| \leq B$}
        \State Observe current frame $I_t$
        \If {\texttt{SELECT}($I_t$) $=$ TRUE}
            \State $\tilde{P}_t \gets \Theta(I_t)$ \Comment{Infer pseudo-labels}
            \State $\mathcal{L}_i \gets \mathcal{L}_i \cup (I_t, \tilde{P}_t) $
        \EndIf
        \State $t \gets t + 1$
    \EndWhile
\EndFor
\Statex

\Comment{\MakeUppercase{Clustering}}
\State $\{E_1,\cdots, E_k\}\gets$ \texttt{CLUSTER}$(\mathcal{L}_1,\cdots, \mathcal{L}_N;K)$
\Statex
\Statex
\Comment{\MakeUppercase{\emph{Students} Fine-Tuning}}

\For{$k = 1, \cdots, K$}
\State $\theta^{E_k} \gets train(\theta, E_k)$
\EndFor

\Statex
\Comment{\MakeUppercase{Update Models}}
\For{$i \in 1,\cdots, N$}
\State  $\theta_i \gets \theta^{E_{map(i)}}$
\EndFor
\State \Return $\{\theta_i\}_{i=1, \cdots, N}$  \Comment{Return updated models}
\end{algorithmic}
\end{algorithm}

This research aims to explore the trade-offs involved in choosing \( K \). On one hand, a small \( K \) implies models trained with a broad diversity and good applicability across various students, potentially with improved generalization. On the other hand, a \( K \) approaching \( N \) may tailor models to the nuances of individual cameras. 

\subsection{Our \texttt{SELECT} Strategy}
\label{sec:selectstrategies}
Our investigation into video sampling strategies underscored the superiority of the \texttt{TOP-CONFIDENCE} approach. In this method, the local DNN \emph{Student} \(\theta\) begins with a warm-up phase by inferring on \(w\) frames, without selecting any for training. For each frame \(I_t\), where \(t \in \{1,\cdots, w\}\), it produces a set of \(L\) bounding boxes \(\tilde{P}_t = \{\tilde{p}_{1t},\cdots,\tilde{p}_{Lt}\}\) along with their associated confidence score \(C_t = \{c_{1t},\cdots,c_{Lt}\}\) predictions, computed as \cite{redmon2016you}. Each frame is then scored \(C(I_t)\) based on the highest confidence score, that is, \(C(I_t) = \max_l c_{lt}\), where \(l \in \{1,..., L\}\). This process establishes a threshold \(\Delta\) as the \((1-\alpha)\)-upper percentile of \(C(I_t)\), aiming for an \(\alpha\) selection rate where \(\mathbb{P}(C(I_t) \geq \Delta) = \alpha\). Subsequently, for $t > w $, a frame $I_t$ is forwarded to the Server if $C(I_t) \geq \Delta$. The pseudo-code is presented in Algorithm \ref{alg:top_conf_detection}.

\begin{algorithm}
\caption{\texttt{TOP-CONFIDENCE} Thresholding for Object Detection}
\label{alg:top_conf_detection}
\begin{algorithmic}[1]
\Require model $\theta$, budget $B$, warm-up period $w$, selection rate $\alpha$.
\Statex
\State Initialize confidence score array $C$ with size $w$
\Statex
\Comment{Collect Confidence Scores}
\For{$t = 1$ to $w$}
    \State $\tilde{P}_t, \tilde{C}_t \gets \theta(I_t)$ \Comment{Infer boxes and scores}
    \State $C[t] \gets \max(\tilde{C}_t)$ \Comment{Max score for $I_t$}
\EndFor
\Statex
\Comment{Compute selection threshold $\Delta$}
\State $\Delta \gets \text{percentile}(C, 1-\alpha)$
\Statex
\Comment{Forward frames meeting $\Delta$ until budget is met}
\State $i \gets 0$ \Comment{Selected frames counter}
\While{$i < B$ and more frames available}
    \State $C_t \gets \max(\theta(I_t))$ \Comment{Get max score for new $I_t$}
    \If{$C_t \geq \Delta$} \Comment{If meets $\Delta$, forward}
        \State Forward $I_t$ to server; $i \gets i + 1$
    \EndIf
    \State $t \gets t + 1$ \Comment{Next frame}
\EndWhile
\end{algorithmic}
\end{algorithm}

Note that \( \alpha \) can be carefully increased to allow more permissive selection, meeting the image budget. This was applied in our experiments (see Section \ref{sec:implementedSELECT}).


\subsection{Clustering}
\label{sec:CLUSTERmethod}
We propose a node clustering method for DNN training. Our method is based on the premise that once a model is fine-tuned on a particular camera domain, it will likely transfer effectively to other camera domains with similar features. Conversely, significant performance drops are expected across dissimilar domains.

\texttt{CLUSTER} uses a \emph{Hierarchical Clustering Algorithm} after the initial training of \( N \) stream-specific \emph{Student} models. This algorithm does not need a pre-specified number of clusters. Instead, the resulting dendrogram can be cut at the appropriate level to obtain an appropriate number of clusters. 

Before discussing our method, we define the validation set \( \mathcal{V}_i \) as the set of image and pseudo-label pairs \((I^{val}_{it}, \tilde{P}_{it})\) from a stream \( \mathcal{X}_i \). We also define the function \( f(\theta, \mathcal{V}) \) measuring the quality of a model \( \theta \) on a set \( \mathcal{V} \).

\textbf{STEP 1 \textendash\; Train \( N \) Stream-specific Models:}
We fine-tune \( N \)\footnote{When \(N \) is too large, we can randomly select a small subset of the streams to conduct this step. This bounds the number of stream-specific models to train, while providing sufficient diversity. The resulting cluster models are then deployed based on their performance on the node's validation set.} stream-specific models, denoted by \( \theta^{\mathcal{L}_i} \). They will subsequently be grouped according to their validation performance on the sets \( \{\mathcal{V}_1, \ldots, \mathcal{V}_N\} \). From an operational standpoint, the choice of \( B \) and number of epochs for this step should align with your specific constraints and use case. Generally, a higher training budget \( B \) allows for better capture of the stream's distribution, thus the transferability (or not) of models will be more pronounced, hence leading to more effective clustering. 
 
\textbf{STEP 2 \textendash \;Compute Cross-Domain Performance:}
We compute the cross-domain performance matrix \( M \) $\in \mathbb{R}^{N\times N}$, with each element \( M_{ij} \) representing the performance metric \( f\left(\theta^{\mathcal{L}_i}; \mathcal{V}_{j}\right) \), which indicates how a model trained on domain $\mathcal{X}_i$ performs on the validation set of $\mathcal{X}_j$. The matrix is defined in Eq.\ref{eq:crossperformancematrix}.

\begin{equation}
\label{eq:crossperformancematrix}
M := \left[ \begin{array}{ccc}
f\left(\theta^{\mathcal{L}_1}; \mathcal{V}_{1}\right) & \cdots & f\left(\theta^{\mathcal{L}_1}; \mathcal{V}_{N}\right) \\ 
\vdots & \ddots & \vdots \\ 
f\left(\theta^{\mathcal{L}_N}; \mathcal{V}_{1}\right) & \cdots & f\left(\theta^{\mathcal{L}_N}; \mathcal{V}_{N}\right)
\end{array} \right]
\end{equation}

\textbf{STEP 3 \textendash \;Compute Distance Matrix:} To quantify the dissimilarities between the models trained on different domains, we calculate the distance matrix \( D \). This matrix is computed using the pairwise Euclidean distances between the models' performance vectors, derived from the cross-domain performance matrix \( M \).

The distance matrix \( D \) is defined as:

\begin{equation}
\label{eq:distancematrix}
D_{ij} = \sqrt{\sum_{k=1}^{N} (M_{ik} - M_{jk})^2} \quad \text{for } i, j = 1, \ldots, N
\end{equation}
\textbf{STEP 4 \textendash \;Apply Single Linkage Clustering:}
Using \cite{SLINK}, we construct a dendrogram, i.e., a tree diagram depicting how clusters merge based on minimum distance. Formally, for two clusters \( A \) and \( B \), the single linkage distance \( L(A, B) \) is given by Eq. \ref{eq:linkage}
\begin{equation}
\label{eq:linkage}
L(A, B) = \min \{ D_{ij} : i \in A, j \in B \}
\end{equation}
where \(D_{ij}\) is defined as in Eq. \ref{eq:distancematrix}. This linkage criterion tends to merge clusters with their nearest elements.

\textbf{STEP 5 \textendash \; Define Clusters:} We define the clusters by selecting a cut-off distance \( t \) on the dendrogram (see Fig. \ref{fig:dendogram}), guided by expertise, statistical methods, or visual analysis.

\section{Materials And Methods}
\label{sec:materialsandmethods}
This section elaborates on the datasets, models, and evaluation metrics employed in our study.

\subsection{Dataset}
\label{sec:datasets}
The Watch and Learn Time-lapse (WALT) dataset \cite{Reddy_2022_CVPR} features footage from nine cameras over several weeks, offering a variety of viewpoints, lighting and weather conditions, including snow and rain. Data collection spans periods ranging from one to four weeks per camera, resulting in a weekly range of 5,000 to 40,000 samples due to asynchronous recordings and differing sampling rates (details can be found in the GitHub repository). Notwithstanding, some instances of burst phenomena are observed, presenting cases of strong temporal redundancy. 

\subsection{Evaluation}
System performance was evaluated by the accuracy of \emph{Student} models on their respective camera.

\subsubsection{Test sets}
We had nine test sets, two from the original paper and seven that we annotated by selecting a week of footage from each camera, which we excluded from training. These sets, now publicly available, facilitate further research (details and links can be found in Supplementary Materials and Github repository).

\subsubsection{Detection accuracy}
In object detection model evaluation, the mAP50-95 metric signifies the mean Average Precision (\textit{mAP}) across various Intersection over Union (\textit{IoU}) thresholds, spanning from 0.50 to 0.95 in increments of 0.05.

\subsubsection{Adjusted Training cost}
We define the training cost \(T\) as the total number of training iterations, which equates to the number of weight updates within a model. To facilitate fair comparisons among models trained on clusters of varying sizes, we introduce an adjusted training cost \(T^{N}\) for a specified number of streams \(N\). Considering the number of samples per stream \(B\), we adjust the epoch count to ensure that the iteration count remains consistent regardless of the cluster count \(K\). This calibration for models trained on any given cluster \(k\), with \(k \in \{1, \ldots, K\}\), is encapsulated in Eq. \ref{eq:constantComplexity}:

\begin{equation}
\label{eq:constantComplexity}
    T^{N}_{k} = \frac{ B }{\text{batch size}} \times \frac{\text{epochs}_{|K=1} \times N}{n_k}
\end{equation}

Here, \(n_k\) represents the stream count in cluster \(k\), \(B\) is the number of images collected per stream, batch size is the number of samples processed in one forward and backward pass, and \(\text{epochs}_{|K=1}\) denotes the number of epochs for the universal model, i.e., when \(K=1\).  

\subsection{Models, Training and Distillation Implementation}
We utilize models pretrained on the COCO dataset \cite{DBLP:journals/corr/LinMBHPRDZ14} to ensure a broad foundational understanding of the object detection task. We employ \textbf{YOLOv8x6\(^{\text{COCO}}\) as the \emph{Teacher} $\Theta^{\text{COCO}}$} and \textbf{YOLOv8n\(^{\text{COCO}}\) as the  \emph{Student}: $\theta^{\text{COCO}}$}. The architectures have 68.2 and 3.2 million parameters, respectively.
Both models serve as the evaluation benchmark. No preprocessing was done, but for post-processing, similar COCO object classes (bikes, cars, motorcycles, buses, and trucks) were grouped into a "vehicle" category.
The \emph{Student} is then fine-tuned by default on 100 epochs, which can vary according to the experiment. The batch size remains constant at 16. All other training parameters are defaults as configured in \cite{yolov8}.  

\subsection{\texttt{SELECT} an Parameters}
\label{sec:implementedSELECT}
Our investigation evaluates, alongside \texttt{Top-Confidence}, three other \texttt{SELECT} strategies:

\begin{itemize}
    \item \texttt{Uniform-Random}:  a frame is chosen if \(s \sim U(0,1) \geq 1-\alpha\). To ensure statistical reliability, we conducted six iterations with six different seeds. The variability in performance is quantified by the Margin of Error (MoE), calculated as \(\text{MoE} = z \times \frac{\sigma}{\sqrt{n}}\), where \(z\) represents the 1.96 z-score correlating with a 95\% confidence level, \(\sigma\) is the standard deviation of the scores, and \(n\) is the effective sample size.
    
    \item \texttt{Least-Confidence}: targets frames with minimal prediction confidence to present the model with challenging instances. This method mirrors the \texttt{Top-Confidence} algorithm (\ref{alg:top_conf_detection}), albeit with a reversed selection criterion, shifting from $C(I_t) \geq \Delta$ to $C(I_t) \leq \Delta$.
    
    \item \texttt{N-First}: opts for the initial \(B\) frames from the stream, establishing an unbiased baseline.
\end{itemize}

Standard parameters are set to $\alpha=0.1$ and $w=720$. Relaxation was done in high-budget scenarios for cameras 3 and 9 due to data constraints, increasing $\alpha$ to 0.55 for camera 9 and 0.25 for camera 3, thus facilitating expedited frame selection to fulfill our image budget.

\section{Results}
\label{sec:experimentsscene-specific}
Section \ref{sec:stream-specific} assesses the efficacy of various \texttt{SELECT} sampling strategies and examines how the number of images per stream (\(B\)) affects the performance of individual \emph{Student} models.

Next, we analyzed the correlation between the complexity of \emph{Teacher} models and their tendency to induce confirmation bias, as detailed in Section \ref{sec:confirmation bias}.

Section \ref{sec:clustersdefinition} considers the clustering of cameras. Sections \ref{sec:clusteringConstantComplexity} and \ref{sec:clusteringConstantMaturity} investigate the influence of cluster count and training complexity on model performance.

\subsection{Impact of \texttt{SELECT} and B}
\label{sec:stream-specific}
\begin{figure*}[htpb]
\centering
\subfloat[\textit{Low} Budget]{
  \includegraphics[width=0.48\linewidth]{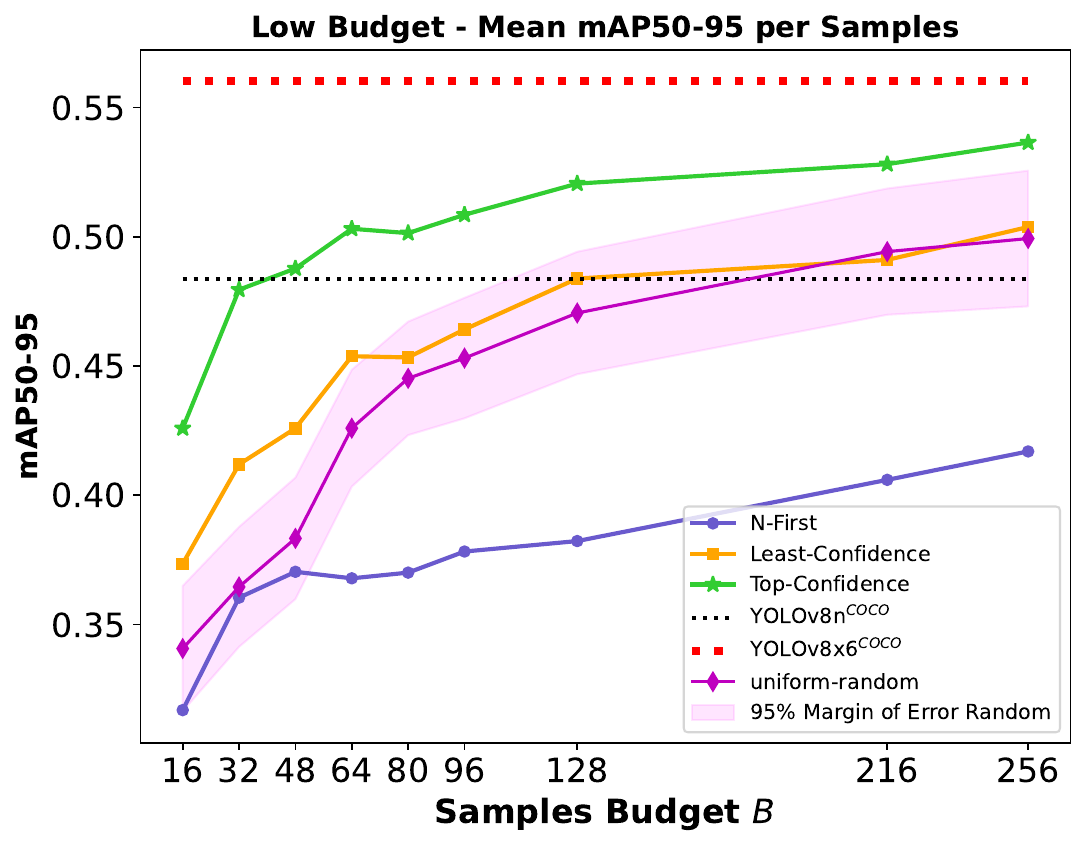}
  \label{fig:sb-week-single}
}
\hfill 
\subfloat[\textit{High} Budget]{
  \includegraphics[width=0.48\linewidth]{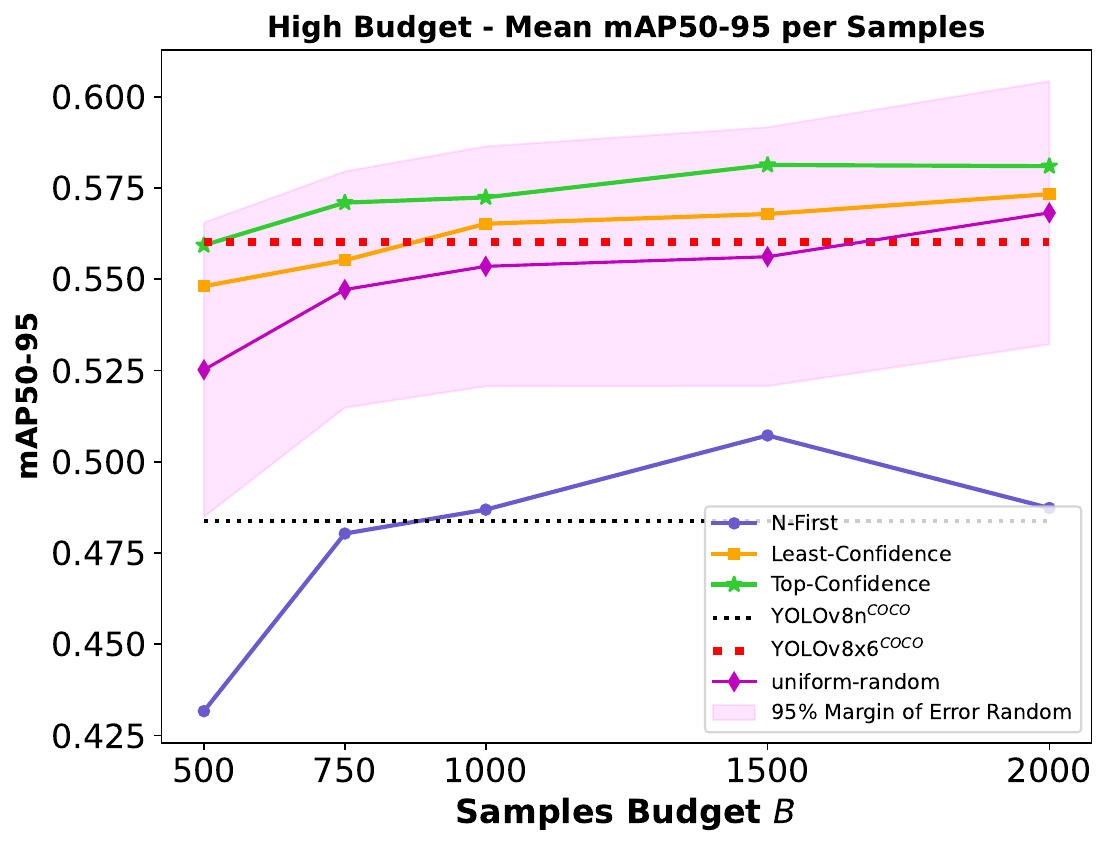}
  \label{fig:sb-month-single}
}
\caption{Average mAP50-95 scores for different training sample sizes, using four SELECT models. The number of epochs is 100. Key observation are: 1) \texttt{Top-Confidence} is the best sampling strategy and 2) fine-tuned compact models can outperform a \emph{Teacher} YOLOv8x6$^{\text{COCO}}$.}
\label{fig:sbstrategies}
\end{figure*}

\emph{Student} \(\theta^{\text{COCO}}\) is fine-tuned using \(B\) samples from a camera, selected on-the-fly according to the \texttt{SELECT} strategies (detailed in Section \ref{sec:implementedSELECT}). The resulting model is then evaluated on the camera's test set. The low budget \(B \leq 256\) experiments were conducted across all cameras for each week, resulting in eighteen pairs per experiment. For the high budget \(B > 500\) scenario, we combined weekly data from each camera to reach the budget count, forming nine pairs per experiment.

Figure \ref{fig:sbstrategies} presents the average mAP50-95 scores as a function of \(B\) across different \texttt{SELECT} strategies, maintaining a consistent 100 epochs for all tests. Key observations include the following:
\begin{enumerate}
    \item \texttt{Top-Confidence} is the best sampling strategy. Conversely, \texttt{N-First} was catastrophic and performed worse than the student when a small $B$ was considered.
    \item Using \texttt{Top-Confidence}, we outperformed the unfine-tuned models $\theta^{\text{COCO}}$ and $\Theta^{\text{COCO}}$ with sample budgets exceeding 48 and 750, respectively.
    \item As $B$ increases, the model quality improves but at a slower rate, suggesting the achievement of peak accuracy.
\end{enumerate}

The findings emphasize the need for careful design of \texttt{SELECT}, particularly when the budget is limited. We observe that when \( B\) is small, a bad selection can severely disrupt learning. Methods such as \texttt{N-First} and \texttt{Random} are \textbf{ill-suited for the redundant nature of video streams and tend to select frames with little or no relevant activity, penalizing the learning process}. Conversely, the \texttt{Top-Confidence} approach \textbf{tends} to select frames containing objects. Furthermore, \textbf{opting for} higher-confidence frames \textbf{can enhance} the accuracy of pseudo-labels in a distillation scheme, while \textbf{selecting the} least confident frames \textbf{tends to amplify} the teacher's inaccuracies. The section \ref{sec:confirmation bias} delves into this phenomenon, known as confirmation bias. 

Although DNNs generally benefit from larger training sets, our observations reveal a plateau, indicating that beyond \(B = 1500\), adding more images from the same stream and acquisition period does not increase model performance further.
\subsection{Confirmation bias}
\label{sec:confirmation bias}
Confirmation bias in semi-supervised or unsupervised learning can lead to significant errors by reinforcing incorrect predictions and thereby misleading learning \cite{arazo2020pseudo}. To identify strategies mitigating it, we conducted two experiments exploring the interplay of \texttt{Least/Top-Confidence} with varying i) \emph{Teacher} and ii) \emph{Student} sizes. Model size (Params) is quantified by the number of parameters in millions. In both experiments, models were trained for 100 epochs.

\subsubsection{Impact of Teacher Size}
In this experiment, \textbf{we manually annotated} \(96\) images sampled by a \(\theta^{\text{COCO}}\) \emph{Student} utilizing the \texttt{Least/Top-Confidence} approach and subsequently compared the performance of these \emph{Students} against scenarios in which a \emph{Teacher} model pseudo-annotated the same set of images. This experiment was conducted across three camera streams\footnote{Namely, Camera 1 from Week 1, Camera 2 from Week 1, and Camera 3 from Week 5. Links to the manual annotations can be found in the GitHub repository.}, employing various model sizes for the \emph{Teachers}. 

\begin{table}[hbtp]
\centering
\caption{mAP50-95 values, and percentage increase per annotator for \texttt{Least/Top-Confidence}. The \emph{Student} is YOLOv8n$^{\text{COCO}}$ with \(B = 96\) samples and 100 epochs.\label{tab:oracles}}
\begin{tabular}{lcccc}
\hline
\textbf{Annotator} & \textbf{Params (M)} &\textbf{Least} & \textbf{Top} & \textbf{\% Increase} \\
\hline
yolov8n & 3.2 & 0.28 & 0.45 & 56\% \\
yolov8s & 11.2 & 0.35 & 0.49 & 37\% \\
yolov8m & 25.9 & 0.37 & 0.52 & 40\% \\
yolov8l & 43.7 & 0.40 & 0.51 & 30\% \\
yolov8x6 & 68.2 & 0.40 & 0.52 & 31\% \\
human & N/A & 0.40 & 0.52 & 27\% \\
\hline
\end{tabular}
\end{table}

The results, detailed in Table \ref{tab:oracles}, reveal that less complex \emph{Teachers} tend to yield inferior \emph{Students}, a situation that the \texttt{Least-Confidence} sampling strategy exacerbates. Intriguingly, human annotations did not offer improvements. This suggests the \texttt{Least-Confidence}'s tendency to choose ambiguous or challenging samples, thereby reinforcing confirmation bias. Conversely, \texttt{Top-Confidence} significantly improves pseudo-label quality and, consequently, model performance, highlighting the critical need for selecting high-quality, clear images for training. Moreover, this finding reiterates the effectiveness of pseudo-labeling, indicating that it nearly matches the performance enhancements provided by human-generated labels.

\subsubsection{Impact of Student Size}
We explored the influence of \emph{Student} model size on performance, utilizing \(B = 256\) images from nine cameras, pseudo-labeled by \(\Theta^{\text{COCO}}\).

Our observations, detailed in Table \ref{tab:studentcomplexity}, yield two key insights, namely, 1) a general enhancement in performance with the increase in \emph{Student} model size, and 2) a growing disparity in the effectiveness of \texttt{Least/Top-Confidence} strategies as the model size expands. These findings indicate that larger models are better equipped to regularize and generalize, a crucial aspect in correcting the inaccuracies inherent in pseudo-labels generated through the \texttt{Least-Confidence} strategy. Furthermore, the \texttt{Top-Confidence} strategy appears to leverage the advanced capabilities of bigger models more efficiently.

\begin{table}[htpb]
\centering
\caption{Comparison of mAP50-95 Values and Percentage Increase per \emph{Student} for Different Strategies. The \emph{Teacher} is YOLOv8x6$^{\text{COCO}}$ and \( B = 256 \) samples and 100 epochs were used.}
\label{tab:studentcomplexity}
\begin{tabular}{lcccc}
\hline
\textbf{Student} & \textbf{Params (M)} & \textbf{Least} & \textbf{Top} & \textbf{\% Increase}\\
\hline
yolov8n & 3.2 & 0.52 & 0.53 & +2.8\% \\
yolov8s & 11.2 & 0.56 & 0.58 & +2.4\% \\
yolov8m & 25.9 & 0.56 & 0.59 & +4.6\% \\
yolov8l & 43.7 & 0.57 & 0.60 & +4.3\% \\
yolov8x & 68.2 & 0.57 & 0.60 & +5.0\% \\
\hline
\end{tabular}
\end{table}

\subsection{Cluster Definition}
\label{sec:clustersdefinition}
\begin{figure*}[htpb]
\centering
\subfloat[Cross-Domain Performance Matrix \( M \).]{\centering\includegraphics[width=0.48\linewidth]{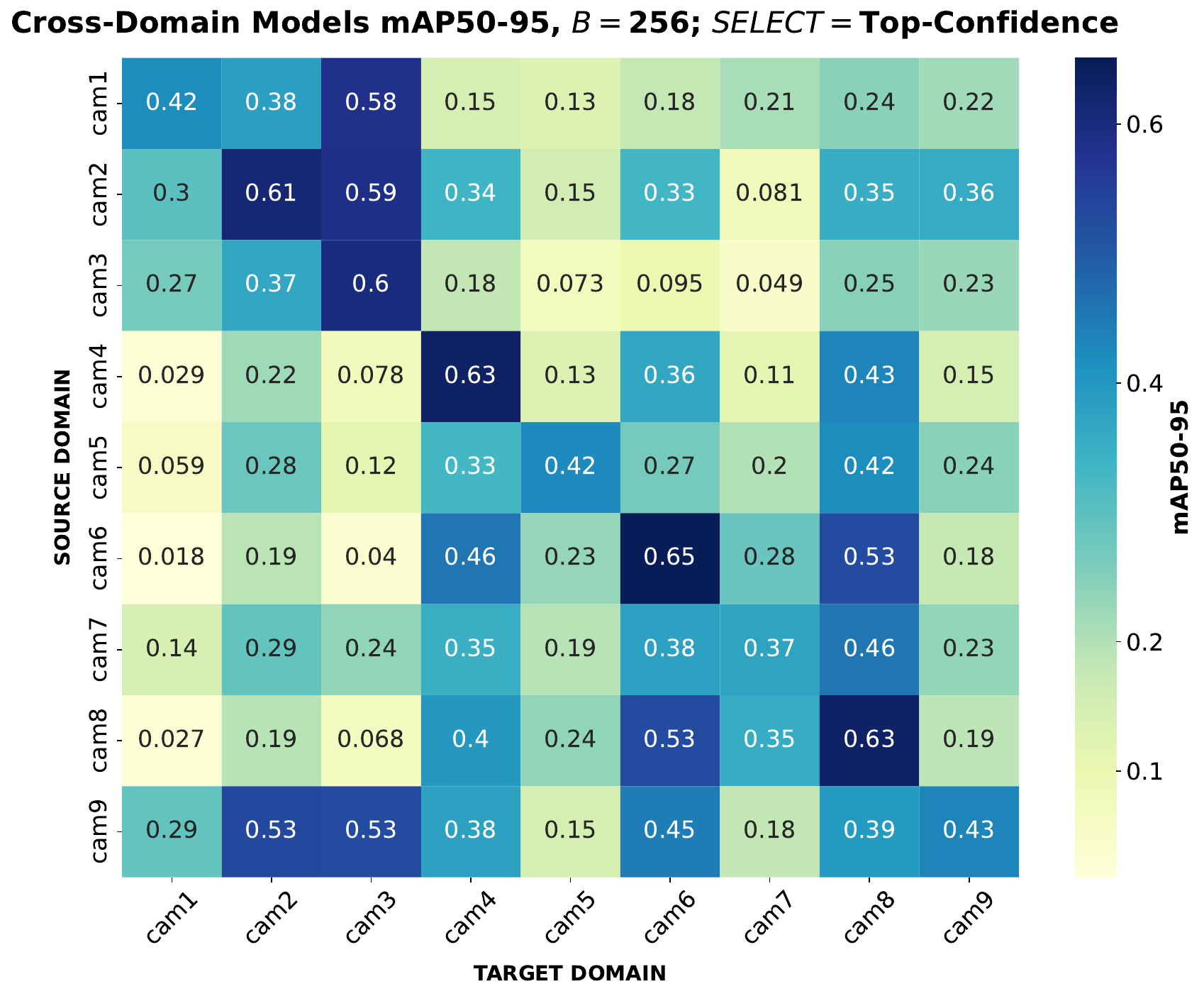}\label{fig:heatmap256threshtopconf}}
\hfill
\subfloat[Dendrogram.]{\centering\includegraphics[width=0.48\linewidth]{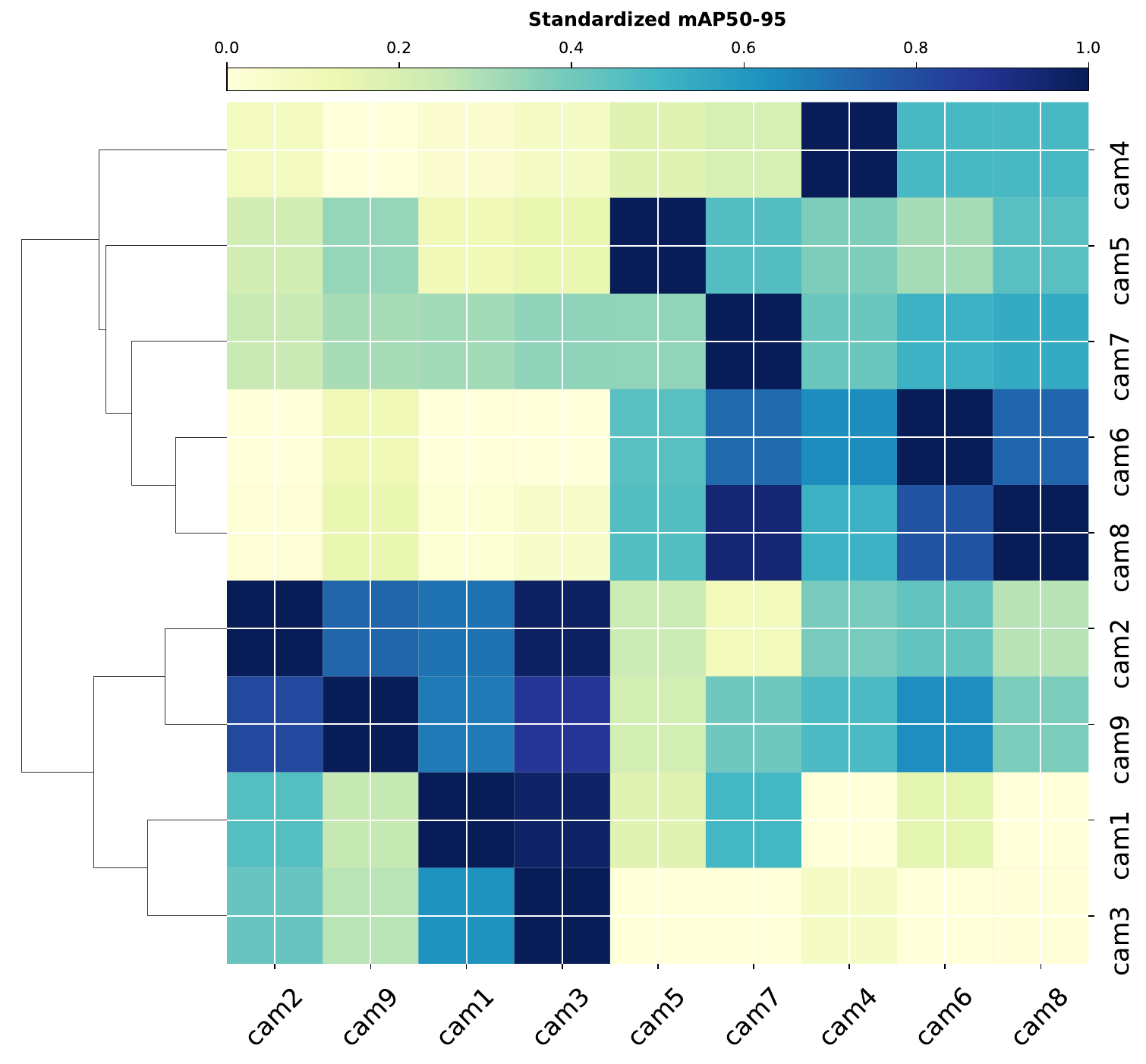}\label{fig:dendogram}}

\caption{Clustering Definition. Fig. \ref{fig:heatmap256threshtopconf} depicts the cross-performance matrix \( M \), where each element \( M_{ij} \), \( i,j=1,\cdots,9 \), represents the mAP50-95 score of a model \( \theta_i \) retrained on source domain \( cam_i \) and evaluated on target domain \( cam_j \). The setting is based on \( B = 256 \) images sampled using \texttt{Top-Confidence}. Fig \ref{fig:dendogram} is the associated dendogram.}
\end{figure*}

The cross-domain performance matrix \( M \) in Figure \ref{fig:heatmap256threshtopconf}, displays the mAP50-95 score of a model \(\theta^{\text{COCO}}\) fine-tuned on source \( cam_i \) and tested on target \( cam_j \), where \( i \) and \( j \) are the nine cameras from the WALT dataset. The analysis utilized the first week's data from each camera, sampling \( B = 256 \) images using \texttt{Top-Confidence}.


Key observations from Figure \ref{fig:heatmap256threshtopconf} are:


\begin{enumerate}
    \item Models perform best within their own domain.
    \item The transfer of models results in performance degradation. The severity is variable. 

\end{enumerate}
The variance in the model's transferability across the stream can be used to group the streams. Having verified our premises, we apply Hierarchical Clustering with \emph{minimum linkage} to detect subtle similarities. Note that in this particular scenario the application of other linkage methods (i.e., \textit{max, average} or \textit{ward}) leads to the same clustering. Cluster definition involved experimenting with various threshold values and observing emerging groups in the dendogram (Figure \ref{fig:dendogram}). The clustering obtained for the nine WALT cameras is presented in Table \ref{tab:clusters} in Appendix.

\subsection{Clustering vs Samples Budget}
\label{sec:clusteringConstantComplexity}
\begin{figure*}[htpb]
\centering
\subfloat[\( B = \{16,32, 48, 64, 80, 96,216,256\} \)]{
  \includegraphics[width=0.48\linewidth]{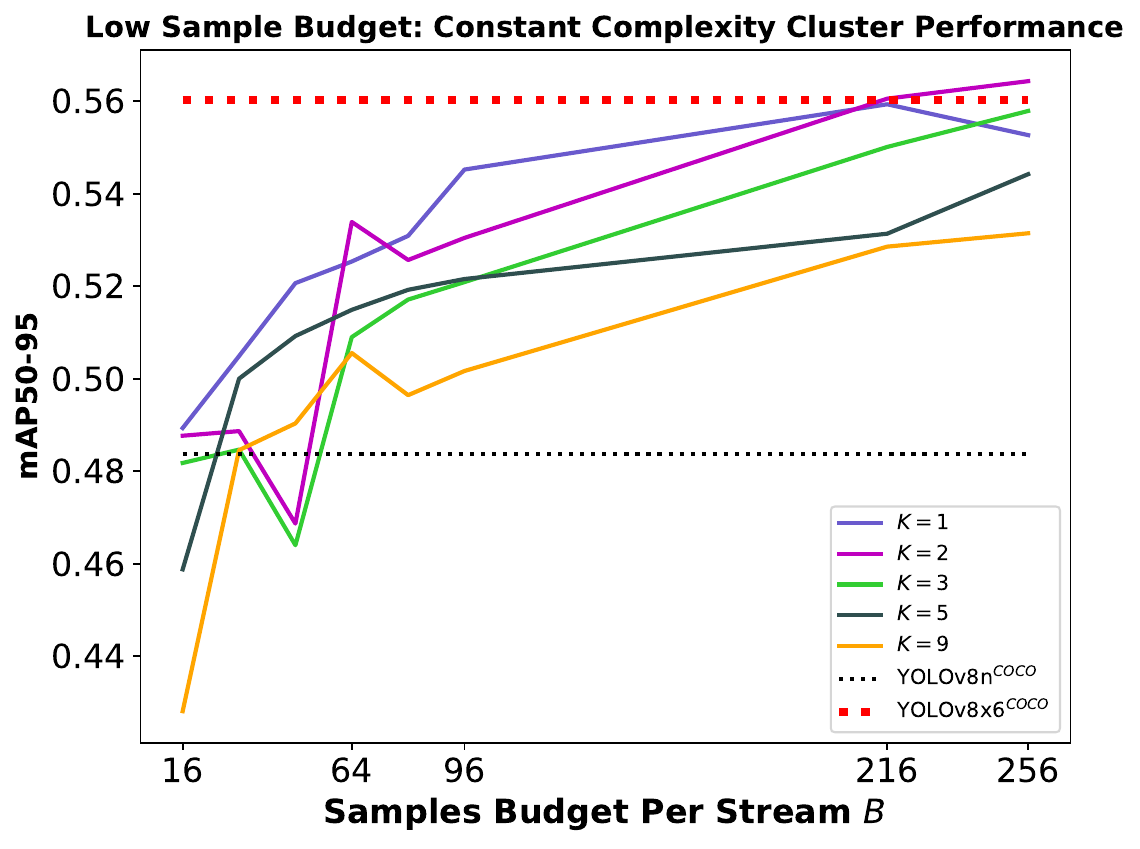}
  \label{fig:sb-week-clustering}
}
\hfill 
\subfloat[\( B = \{500, 750, 1000, 1500, 2000\} \)]{
  \includegraphics[width=0.48\linewidth]{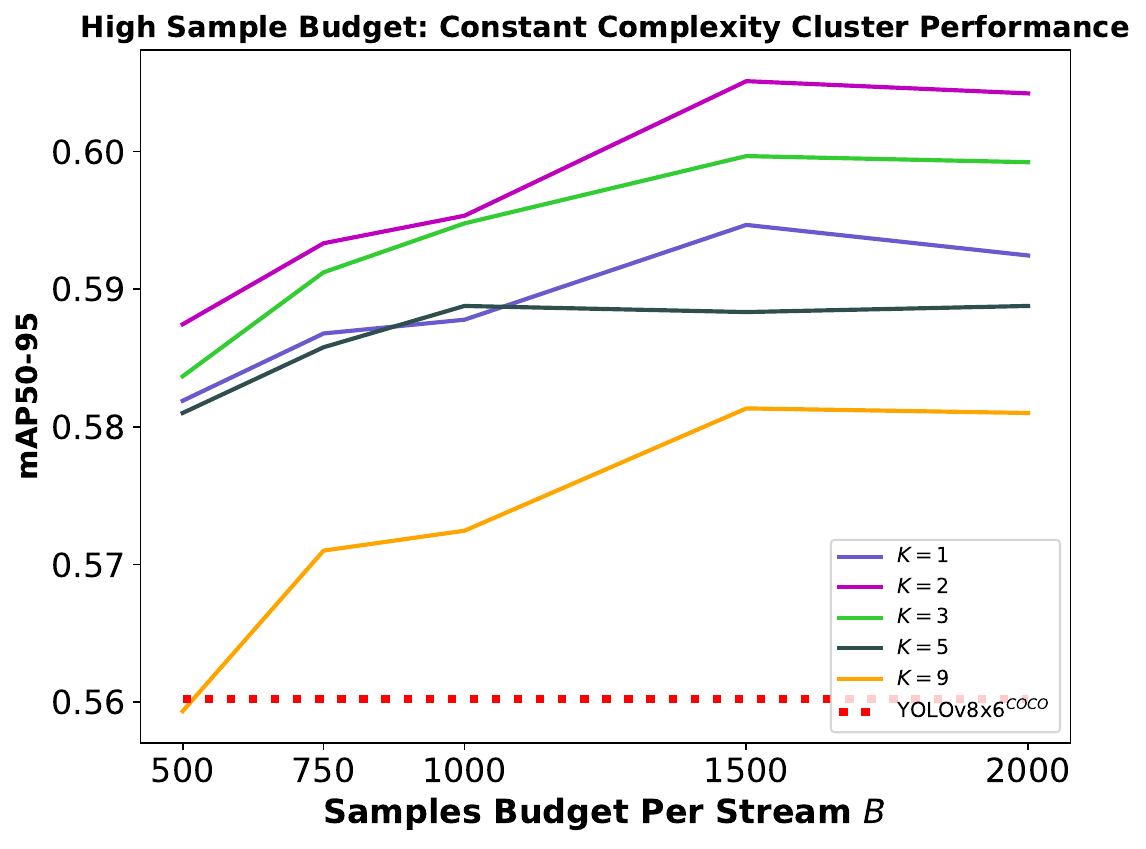}
  \label{fig:sb-months-clustering}
}
\caption{Mean mAP50-95 scores per sample budget per stream (\(B\)) for varying numbers of clusters were observed. Models underwent training for 100 epochs with a batch size of 16. Key findings indicate that, at a constant complexity, a universal model (\(K=1\)) is preferable for lower \(B\) values. However, for larger \(B\) values, segmenting the system into two or three clusters yields superior outcomes.}
\label{fig:clusteringresults}
\end{figure*}

Figure \ref{fig:clusteringresults} presents mAP50-95 scores as a function of the amount of images per stream \( B \) for different numbers of clusters \( K \). The clusters' definition follows Fig. \ref{fig:dendogram}. The observations are as follow:  

\begin{enumerate}
    \item On smaller budgets, training one model for all streams is superior. In fact, the smaller $ K $ is, the better are the models. 
    \item On larger budgets, $K = 2$ and $ K = 3 $ are dominating while the scenario $K = N = 9$ is underperforming. 
    \item For all $ K \in (1,\; \cdots \;,9) $, the models reach their peak accuracy around $ B = 1500$ samples per stream equivalent to total amount of iterations \(T = 84 375\), irrespective of \( K \).
\end{enumerate}

The amount of images per stream \( B \) dictates the best \( K \). Training a single model for all streams proves most effective for a low sample budget \( B \), reflecting the data-hungry nature of DNN. However, for a larger \( B \), clustering increases performance. In fact, there is a a ``sweet spot'' between stream-specific and universal models. This suggest that clustering offers robustness by leveraging more samples and a more specific distribution facilitating the learning, particularly in less complex architectures. Yet, there's a potential downside to excessive clustering, as it might overlook the need for a model of diversity. 

\subsection{Clustering vs Constant Iteration per Model}
Since a larger K induces more models trained with smaller training sets, we also investigated the relationship between \(K \) and the number of training epochs. 

\begin{figure*}[htpb]
    \includegraphics[width=\linewidth]{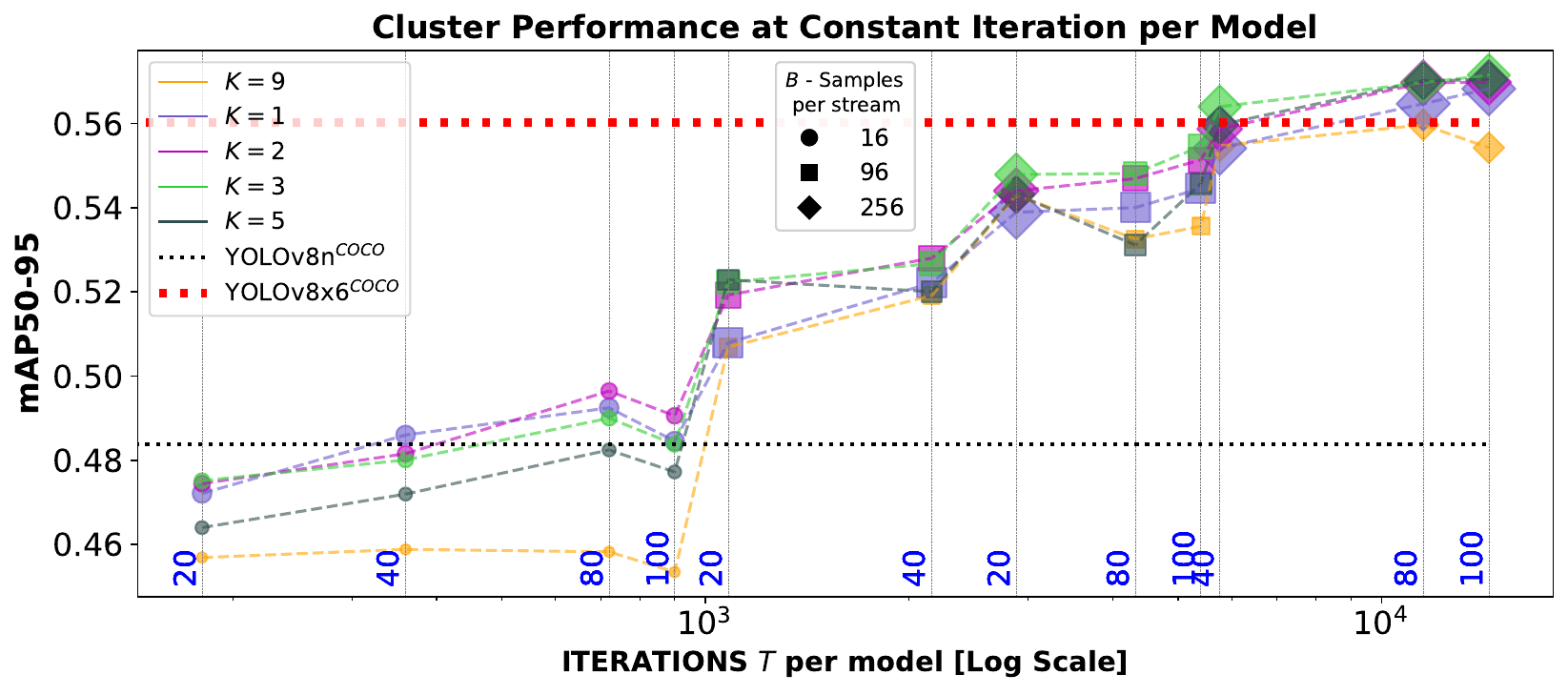}
    \caption{Mean mAP50-95 scores are presented over log-scaled iterations (\(T\)) for each model. Markers denote sample sizes per stream (\(B = 16, 96, 256\)). Vertical lines indicate the epochs for \(K = 1\). For \(K > 1\), the epoch count is adjusted to maintain constant iteration counts across models, following Equation \ref{eq:constantComplexity}. Our analysis reveals that an increase in epoch counts benefits smaller cluster configurations (\(K= 3, 5, 9\)), enabling them to achieve comparable or superior performance to more universal configurations (\(K = 1, 2\)).\label{fig:maturity}}
\end{figure*}

\label{sec:clusteringConstantMaturity}
Figure \ref{fig:maturity} illustrates the mAP50-95 scores as a function of the number of iterations \(T^{K=1}_{1}\) in a log-scale for the universal model. Models for clusters with \(K > 1\) adjust their epoch counts, in accordance with Eq. \ref{eq:constantComplexity}, to maintain a consistent number of training iterations (i.e., model updates) across all models, regardless of \(K\). The figure uses distinct markers to represent sample sizes of \(B = 16, \;96\) and \(256\) per stream, and vertical lines to mark the epoch counts when \(K = 1\). 


The key observations are: 

\begin{enumerate}
    \item Increasing \( B \) from 16 to 96, and further to 256, results in substantial improvements outpacing gains from extended epoch counts.
    \item In comparison with Figure \ref{fig:clusteringresults}, smaller clusters (\( K = 5, 9 \)) bridged the performance gap with larger-cluster scenarios, particularly as iteration counts increased. 
    \item Performance tends to plateau and even decline after reaching a base epoch (epochs$_{K=1}$) of 80 epochs, suggesting the onset of overfitting beyond a base 100 epochs.
\end{enumerate}

We conclude that careful escalation of the epochs enabled smaller \( K \) values to optimize model performance while minimizing the risk of overfitting.

\section{Discussion}
\label{sec:discussion}
\subsection{Insights}
The CSBAD framework performance is determined by the number of images per stream ($B$), active learning strategies (\texttt{SELECT}), number of clusters ($K$), number of epochs, and dimensions of the \emph{Students} along with the complexity of the \emph{Teacher}. 

Our analysis confirms the expected significance of $B$
. We found that satisfactory results could be achieved with $B>64$, with performance gains plateauing beyond $B\geq1500$. The epoch count, while relevant, proved to be of secondary importance (see Fig. \ref{fig:maturity}). 
Increasing \(B\) means more images are used for training, potentially boosting performance but requiring more resources. Naturally, the success of the \texttt{Top-Confidence} strategy highlights the importance of selecting clear, well-lit images with identifiable instances for training in order to minimize the budget. 

As for the optimal number of clusters ($K$), there is no one-size-fits-all answer. The choice depends on the system size ($N$), available hardware, and budget \( B \) constraints. A smaller $B$ or $N$ requires a smaller \( K \) to create larger clusters, providing sufficient samples and training iterations for model training. Conversely, a larger $N$, $B$, or number of iterations per model \( T \) shifts the constraint to hardware capabilities, encouraging model tailoring. Essentially, $K$ \textbf{acts as an ``adjusting variable" within the system} to achieve both a \textbf{specificity-diversity trade-off} and sufficient \textbf{training iterations per model}. 

Finally, the superior performance of compact fine-tuned \emph{Students} over a large general-purpose \emph{Teacher} demonstrates the distillation's effectiveness, simultaneously reducing complexity and boosting domain-specific performance without the need for a human annotator. Regarding model size, we observe that larger \emph{Teacher} or \emph{Student} models generally yield better outcomes.

\subsection{Limitations and perspectives}
\subsubsection{Scalability}
Inherently, large-scale systems exhibit irregularity and unpredictability. Therefore, the applicability of current \( K \) results is limited to comparable scales, and determining system behaviors for higher order systems (e.g., \( N > 100 \)) requires experimentation at much higher scales. 

\subsubsection{Continuous Deployment}
Sustaining a budget \( b \), where \( b \leq B \), for each \( \mathcal{L}_i \) through successive deployment cycles (i.e., integration of a new device or regular updates), can have advantages. First, this strategy saves resources, as only $B-b$ new images and annotations are required in the next deployment cycle. Second, this approach could mitigate \emph{catastrophic forgetting} \cite{catastrophicforgettingfrench1999}, a scenario where models perform well in new classes but decline in older ones \cite{meta-learningfewshot, catastrophicforgettingfrench1999}\footnote{The causes behind catastrophic forgetting include (1) network drift, where the neural network’s superior data fitting ability leads it to drift quickly from the feature space learned from the old class training data to that of the new class data and (2) inter-class confusion, where the boundaries between new and old classes are not well established because they have never been trained on together \cite{semantincknowledguideclassincrementallearning}.}.  As this require further assumptions on the frequency of fine-tuning and the use case, we considered \( b = 0\) in our experiments. The extension can be done by changing Line 3 in Algorithm \ref{alg:1} to  $\mathcal{L}_i \gets \text{\texttt{RECYCLE}}(\mathcal{L}_i)$, where \texttt{RECYCLE} manages the storage of $\{\mathcal{L}_i\}_{i=1,...,N}$ for subsequent deployment.

\subsubsection{Non-optimal Resource Management}
Our approach necessitates initiating model training from a general-purpose model with each iteration of the framework, leading to suboptimal resource utilization. This process neglects the potential advantages of leveraging previously trained models. For instance, in the \texttt{CLUSTER} approach, although \(N\) stream-specific models are initially developed, they are not utilized in subsequent training for cluster-specific models when \(K < N\), resulting in their underutilization.

Future research should investigate \textbf{incremental deployment}, with the effective utilization of pretrained models, potentially through methods like weight aggregation \cite{aggregationweights} and fine-tuning for fewer epochs, to boost learning efficiency and reduce training costs.

\section{Conclusions}
We developed a scalable framework aimed at streamlining model deployment across various video streams. This framework includes collecting video stream samples, utilizing general-purpose models for pseudo-labelling, and performing data clustering to enhance retraining efficiency. When it is applied to object detection in CCTV streams, CSBAD yields  significant improvements. The \texttt{Top-Confidence} SELECT function, focusing on high-confidence samples from compact DNN models, significantly increases model accuracy with minimal labeling effort. This method addresses the challenges associated with active distillation, particularly in minimizing pseudo-label inaccuracies. The introduction of \(K\), a variable representing the number of clusters, is crucial for adapting CSBAD to diverse system configurations, emphasizing the importance of balancing specificity, diversity, sample volume, and training epochs to boost the framework's flexibility. In this sense, CSBAD advocates for the clustering of camera domains to move beyond the constraints of specialized or universal networks, enhancing adaptability and performance.

%



\label{sec:conlcusion}

\section*{Appendix - Clustering Table for WALT}
\label{sec:appendixTable}
\begin{table}[htpb]
    \caption{Clusters\label{tab:clusters} based on cutting the dendogram for different thresholds. 
    }
    \centering
    \small 
    \begin{tabular}{cccc}
        \toprule
        \textbf{Thresh.} & \textbf{\( K \)} & \textbf{Clust. No.} & \textbf{Cams} \\ 
        \midrule
        1.2 & 2 & 1 & 1, 2, 3, 9 \\
        & & 2 & 4, 5, 6, 7, 8 \\
        \hline
        \addlinespace
        1.05 & 3 & 1 & 4, 5, 6, 7, 8 \\
        & & 2 & 1, 3 \\
        & & 3 & 2, 9 \\
        \hline
        \addlinespace
        0.95 & 4 & 1 & 5, 6, 7, 8 \\
        & & 2 & 1, 3 \\
        & & 3 & 2, 9 \\
        & & 4 & 4 \\
        \hline
        \addlinespace
        $< 1$ & 5 & 1 & 6, 7, 8 \\
        & & 2 & 4 \\
        & & 3 & 2, 9 \\
        & & 4 & 5 \\
        & & 5 & 1, 3 \\
        \bottomrule
    \end{tabular}
\end{table}

\section*{Acknowledgments}
This work was partially funded by Win2WAL (\#1910045) and Trusted AI Labs. We also thank \emph{Openhub} for providing the equipment. We would also like to express our gratitude to Charlotte Nachtegael for her initial work on the relation between model size and accuracy. We also thanks Stéphane Galland for the discussion.


%



\bibliographystyle{unsrtnat}
\bibliography{ref}

\begin{thebibliography}{54}
\providecommand{\natexlab}[1]{#1}
\providecommand{\url}[1]{\texttt{#1}}
\expandafter\ifx\csname urlstyle\endcsname\relax
  \providecommand{\doi}[1]{doi: #1}\else
  \providecommand{\doi}{doi: \begingroup \urlstyle{rm}\Url}\fi

\bibitem[Beyer et~al.(2022)Beyer, Zhai, Royer, Markeeva, Anil, and Kolesnikov]{LucasBeyer2022}
Lucas Beyer, Xiaohua Zhai, Am\'elie Royer, Larisa Markeeva, Rohan Anil, and Alexander Kolesnikov.
\newblock Knowledge distillation: A good teacher is patient and consistent.
\newblock In \emph{Proceedings of the IEEE/CVF Conference on Computer Vision and Pattern Recognition (CVPR)}, pages 10925--10934, June 2022.

\bibitem[Manjah et~al.(2023)Manjah, Cacciarelli, Standaert, Benkedadra, de~Hertaing, Macq, Galland, and De~Vleeschouwer]{sbad}
Dani Manjah, Davide Cacciarelli, Baptiste Standaert, Mohamed Benkedadra, Gauthier~Rotsart de~Hertaing, Beno{\^\i}t Macq, St\'ephane Galland, and Christophe De~Vleeschouwer.
\newblock Stream-based active distillation for scalable model deployment.
\newblock In \emph{Proceedings of the IEEE/CVF Conference on Computer Vision and Pattern Recognition (CVPR) Workshops}, pages 4998--5006, June 2023.

\bibitem[Hill(1990)]{whatisscal}
Mark~D. Hill.
\newblock What is scalability?
\newblock \emph{SIGARCH Comput. Archit. News}, 18\penalty0 (4):\penalty0 18–21, dec 1990.
\newblock ISSN 0163-5964.

\bibitem[Hossfeld et~al.(2023)Hossfeld, Heegaard, and Kellerer]{recentscalabi}
Tobias Hossfeld, Poul~E. Heegaard, and Wolfgang Kellerer.
\newblock Comparing the scalability of communication networks and systems.
\newblock \emph{IEEE Access}, pages 1--1, 2023.
\newblock \doi{10.1109/ACCESS.2023.3314201}.

\bibitem[Fox et~al.(1997)Fox, Gribble, Chawathe, Brewer, and Gauthier]{clusterscalabledef}
Armando Fox, Steven~D. Gribble, Yatin Chawathe, Eric~A. Brewer, and Paul Gauthier.
\newblock Cluster-based scalable network services.
\newblock In \emph{Proceedings of the Sixteenth ACM Symposium on Operating Systems Principles}, SOSP '97, page 78–91, New York, NY, USA, 1997. Association for Computing Machinery.
\newblock ISBN 0897919165.

\bibitem[Hinton et~al.(2015)Hinton, Vinyals, and Dean]{hinton2015distilling}
Geoffrey Hinton, Oriol Vinyals, and Jeff Dean.
\newblock Distilling the knowledge in a neural network.
\newblock \emph{NIPS 2014 Deep Learning Workshop}, 2015.

\bibitem[Cheng et~al.(2018)Cheng, Wang, Zhou, and Zhang]{cheng2017survey}
Yu~Cheng, Duo Wang, Pan Zhou, and Tao Zhang.
\newblock Model compression and acceleration for deep neural networks: The principles, progress, and challenges.
\newblock \emph{IEEE Signal Processing Magazine}, 2018.

\bibitem[Mishra and Gupta(2023)]{mishra2023designing}
Rahul Mishra and Hari~Prabhat Gupta.
\newblock Designing and training of lightweight neural networks on edge devices using early halting in knowledge distillation.
\newblock \emph{IEEE Transactions on Mobile Computing}, 2023.

\bibitem[Tanghatari et~al.(2023)Tanghatari, Kamal, Afzali-Kusha, and Pedram]{tanghatari2023federated}
Ehsan Tanghatari, Mehdi Kamal, Ali Afzali-Kusha, and Massoud Pedram.
\newblock Federated learning by employing knowledge distillation on edge devices with limited hardware resources.
\newblock \emph{Neurocomputing}, 531:\penalty0 87--99, 2023.

\bibitem[Saito et~al.(2019)Saito, Watanabe, Ushiku, and Harada]{saito2019semi}
Kuniaki Saito, Kohei Watanabe, Yoshitaka Ushiku, and Tatsuya Harada.
\newblock Semi-supervised learning for domain adaptation.
\newblock In \emph{Proceedings of the AAAI Conference on Artificial Intelligence}, volume~33, 2019.

\bibitem[Cioppa et~al.(2019)Cioppa, Deliege, Istasse, De~Vleeschouwer, and Van~Droogenbroeck]{cioppa2019arthus}
Anthony Cioppa, Adrien Deliege, Maxime Istasse, Christophe De~Vleeschouwer, and Marc Van~Droogenbroeck.
\newblock Arthus: Adaptive real-time human segmentation in sports through online distillation.
\newblock In \emph{Proceedings of the IEEE/CVF Conference on Computer Vision and Pattern Recognition Workshops}, pages 0--0, 2019.

\bibitem[Mullapudi et~al.(2019)Mullapudi, Chen, Zhang, Ramanan, and Fatahalian]{mullapudi2019online}
Ravi~Teja Mullapudi, Steven Chen, Keyi Zhang, Deva Ramanan, and Kayvon Fatahalian.
\newblock Online model distillation for efficient video inference.
\newblock In \emph{Proceedings of the IEEE/CVF International Conference on Computer Vision}, pages 3573--3582, 2019.

\bibitem[Rivas et~al.(2022)Rivas, Guim, Polo, Silva, Berral, and Carrera]{rivas2022towards}
Daniel Rivas, Francesc Guim, Jord{\`a} Polo, Pubudu~M Silva, Josep~Ll Berral, and David Carrera.
\newblock Towards automatic model specialization for edge video analytics.
\newblock \emph{Future Generation Computer Systems}, 2022.

\bibitem[Habibian et~al.(2022)Habibian, Ben~Yahia, Abati, Gavves, and Porikli]{delta_distillation_2022}
Amirhossein Habibian, Haitam Ben~Yahia, Davide Abati, Efstratios Gavves, and Fatih Porikli.
\newblock Delta distillation for efficient video processing.
\newblock In Shai Avidan, Gabriel Brostow, Moustapha Ciss{\'e}, Giovanni~Maria Farinella, and Tal Hassner, editors, \emph{Computer Vision -- ECCV 2022}, pages 213--229, Cham, 2022. Springer Nature Switzerland.
\newblock ISBN 978-3-031-19833-5.

\bibitem[Vilouras et~al.(2023)Vilouras, Liu, Sanchez, O’Neil, and Tsaftaris]{vilouras2023group}
Konstantinos Vilouras, Xiao Liu, Pedro Sanchez, Alison~Q O’Neil, and Sotirios~A Tsaftaris.
\newblock Group distributionally robust knowledge distillation.
\newblock In \emph{International Workshop on Machine Learning in Medical Imaging}, pages 234--242. Springer, 2023.

\bibitem[Baykal et~al.(2022)Baykal, Trinh, Iliopoulos, Menghani, and Vee]{baykal2022robust}
Cenk Baykal, Khoa Trinh, Fotis Iliopoulos, Gaurav Menghani, and Erik Vee.
\newblock Robust active distillation.
\newblock \emph{arXiv preprint arXiv:2210.01213}, 2022.

\bibitem[Goldblum et~al.(2020)Goldblum, Fowl, Feizi, and Goldstein]{goldblum2020adversarially}
Micah Goldblum, Liam Fowl, Soheil Feizi, and Tom Goldstein.
\newblock Adversarially robust distillation.
\newblock In \emph{Proceedings of the AAAI Conference on Artificial Intelligence}, volume~34, pages 3996--4003, 2020.

\bibitem[Settles(2009)]{Settles2009}
Burr Settles.
\newblock Active learning literature survey.
\newblock \emph{Computer Sciences Technical Report, University of Wisconsin–Madison}, 2009.

\bibitem[Sener and Savarese(2017)]{Sener2017}
Ozan Sener and Silvio Savarese.
\newblock Active learning for convolutional neural networks: A core-set approach.
\newblock In \emph{ICLR 2018}, 8 2017.

\bibitem[Yoo and Kweon(2019)]{Yoo2019}
Donggeun Yoo and In~So Kweon.
\newblock Learning loss for active learning.
\newblock In \emph{Proceedings of the IEEE/CVF Conference on Computer Vision and Pattern Recognition (CVPR)}, June 2019.

\bibitem[Ash et~al.(2019)Ash, Zhang, Krishnamurthy, Langford, and Agarwal]{Ash2019}
Jordan~T. Ash, Chicheng Zhang, Akshay Krishnamurthy, John Langford, and Alekh Agarwal.
\newblock Deep batch active learning by diverse, uncertain gradient lower bounds.
\newblock In \emph{2020 International Conference on Learning Representations}, 6 2019.

\bibitem[Elhamifar et~al.(2013)Elhamifar, Sapiro, Yang, and Sasrty]{Elhamifar2013}
Ehsan Elhamifar, Guillermo Sapiro, Allen Yang, and S.~Shankar Sasrty.
\newblock A convex optimization framework for active learning.
\newblock In \emph{Proceedings of the IEEE International Conference on Computer Vision}, pages 209--216. Institute of Electrical and Electronics Engineers Inc., 2013.
\newblock ISBN 9781479928392.
\newblock \doi{10.1109/ICCV.2013.33}.

\bibitem[Agarwal et~al.(2020)Agarwal, Arora, Anand, and Arora]{Agarwal2020}
Sharat Agarwal, Himanshu Arora, Saket Anand, and Chetan Arora.
\newblock Contextual diversity for active learning.
\newblock In \emph{European Conference on Computer Vision (ECCV) 2020}, 8 2020.

\bibitem[Prabhu et~al.(2020)Prabhu, Chandrasekaran, Saenko, and Hoffman]{Prabhu}
Viraj Prabhu, Arjun Chandrasekaran, Kate Saenko, and Judy Hoffman.
\newblock Active domain adaptation via clustering uncertainty-weighted embeddings.
\newblock In \emph{International Conference on Computer Vision (ICCV) 2021}, 2020.

\bibitem[Sinha et~al.(2019)Sinha, Ebrahimi, and Darrell]{sinha2019variational}
Samarth Sinha, Sayna Ebrahimi, and Trevor Darrell.
\newblock Variational adversarial active learning.
\newblock In \emph{Proceedings of the IEEE/CVF International Conference on Computer Vision}, pages 5972--5981, 2019.

\bibitem[Yuan et~al.(2021)Yuan, Wan, Fu, Liu, Xu, Ji, and Ye]{Yuan2021}
Tianning Yuan, Fang Wan, Mengying Fu, Jianzhuang Liu, Songcen Xu, Xiangyang Ji, and Qixiang Ye.
\newblock Multiple instance active learning for object detection.
\newblock In \emph{IEEE Conference on Computer Vision and Pattern Recognition (CVPR)}, 4 2021.
\newblock URL \url{http://arxiv.org/abs/2104.02324}.

\bibitem[Cacciarelli et~al.(2022{\natexlab{a}})Cacciarelli, Kulahci, and Tyssedal]{SBAL}
Davide Cacciarelli, Murat Kulahci, and John~Sølve Tyssedal.
\newblock Stream-based active learning with linear models.
\newblock \emph{Knowledge-Based Systems}, 254:\penalty0 109664, 10 2022{\natexlab{a}}.
\newblock ISSN 09507051.
\newblock \doi{10.1016/j.knosys.2022.109664}.

\bibitem[Cacciarelli et~al.(2023)Cacciarelli, Kulahci, and Tyssedal]{ROAL}
Davide Cacciarelli, Murat Kulahci, and John~Sølve Tyssedal.
\newblock Robust online active learning.
\newblock \emph{Quality and Reliability Engineering International}, ENBIS Special Issue, 2023.
\newblock \doi{https://doi.org/10.1002/qre.3392}.
\newblock URL \url{https://onlinelibrary.wiley.com/doi/abs/10.1002/qre.3392}.

\bibitem[Cacciarelli et~al.(2022{\natexlab{b}})Cacciarelli, Kulahci, and Tyssedal]{ICML_SBAL}
Davide Cacciarelli, Murat Kulahci, and John~Sølve Tyssedal.
\newblock Online active learning for soft sensor development using semi-supervised autoencoders.
\newblock In \emph{ICML 2022 Workshop on Adaptive Experimental Design and Active Learning in the Real World}, 2022{\natexlab{b}}.
\newblock \doi{10.48550/arXiv.2212.13067}.
\newblock URL \url{https://arxiv.org/abs/2212.13067}.

\bibitem[Rožanec et~al.(2022)Rožanec, Trajkova, Dam, Fortuna, and Mladenić]{ROZANEC2022277}
Jože~M. Rožanec, Elena Trajkova, Paulien Dam, Blaž Fortuna, and Dunja Mladenić.
\newblock Streaming machine learning and online active learning for automated visual inspection.
\newblock \emph{IFAC-PapersOnLine}, 55\penalty0 (2):\penalty0 277--282, 2022.
\newblock ISSN 2405-8963.
\newblock \doi{https://doi.org/10.1016/j.ifacol.2022.04.206}.
\newblock URL \url{https://www.sciencedirect.com/science/article/pii/S2405896322002075}.
\newblock 14th IFAC Workshop on Intelligent Manufacturing Systems IMS 2022.

\bibitem[Narr et~al.(2016)Narr, Triebel, and Cremers]{Narr2016}
Alexander Narr, Rudolph Triebel, and Daniel Cremers.
\newblock Stream-based active learning for efficient and adaptive classification of 3d objects.
\newblock In \emph{Proceedings - IEEE International Conference on Robotics and Automation}, volume 2016-June, pages 227--233. Institute of Electrical and Electronics Engineers Inc., 6 2016.
\newblock ISBN 9781467380263.
\newblock \doi{10.1109/ICRA.2016.7487138}.

\bibitem[Aner and Kender(2002)]{aner2002video}
Aya Aner and John~R Kender.
\newblock Video summaries through mosaic-based shot and scene clustering.
\newblock In \emph{Computer Vision—ECCV 2002: 7th European Conference on Computer Vision Copenhagen, Denmark, May 28--31, 2002 Proceedings, Part IV 7}, pages 388--402. Springer, 2002.

\bibitem[Sun and Gong(2023)]{sun2023hierarchical}
Shengyang Sun and Xiaojin Gong.
\newblock Hierarchical semantic contrast for scene-aware video anomaly detection.
\newblock In \emph{Proceedings of the IEEE/CVF Conference on Computer Vision and Pattern Recognition}, pages 22846--22856, 2023.

\bibitem[Wang et~al.(2013)Wang, Wang, and Li]{wang2013intelligent}
Yan Wang, Hengyu Wang, and Xirui Li.
\newblock An intelligent recommendation system model based on style for virtual home furnishing in three-dimensional scene.
\newblock In \emph{2013 International Symposium on Computational and Business Intelligence}, pages 213--216. IEEE, 2013.

\bibitem[Li et~al.(2023)Li, Wang, Li, Wei, Shi, Ling, Chen, Liu, Li, and Zheng]{li2023hierarchical}
Zongyi Li, Runsheng Wang, He~Li, Bohao Wei, Yuxuan Shi, Hefei Ling, Jiazhong Chen, Boyuan Liu, Zhongyang Li, and Hanqing Zheng.
\newblock Hierarchical clustering and refinement for generalized multi-camera person tracking.
\newblock In \emph{Proceedings of the IEEE/CVF Conference on Computer Vision and Pattern Recognition}, pages 5519--5528, 2023.

\bibitem[Sz{\H{u}}cs et~al.(2023)Sz{\H{u}}cs, Borsodi, and Papp]{szHucs2023multi}
G{\'a}bor Sz{\H{u}}cs, Reg{\H{o}} Borsodi, and D{\'a}vid Papp.
\newblock Multi-camera trajectory matching based on hierarchical clustering and constraints.
\newblock \emph{Multimedia Tools and Applications}, pages 1--24, 2023.

\bibitem[Wang et~al.(2008)Wang, Tieu, and Grimson]{wang2008correspondence}
Xiaogang Wang, Kinh Tieu, and W~Eric~L Grimson.
\newblock Correspondence-free multi-camera activity analysis and scene modeling.
\newblock In \emph{2008 IEEE Conference on Computer Vision and Pattern Recognition}, pages 1--8. IEEE, 2008.

\bibitem[Specker et~al.(2021)Specker, Stadler, Florin, and Beyerer]{specker2021occlusion}
Andreas Specker, Daniel Stadler, Lucas Florin, and Jurgen Beyerer.
\newblock An occlusion-aware multi-target multi-camera tracking system.
\newblock In \emph{Proceedings of the IEEE/CVF conference on computer vision and pattern recognition}, pages 4173--4182, 2021.

\bibitem[Patel et~al.(2021)Patel, Yao, Qiang, Ooi, and Zimmermann]{patel2021multi}
Toshal Patel, Alvin Yan~Hong Yao, Yu~Qiang, Wei~Tsang Ooi, and Roger Zimmermann.
\newblock Multi-camera video scene graphs for surveillance videos indexing and retrieval.
\newblock In \emph{2021 IEEE International Conference on Image Processing (ICIP)}, pages 2383--2387. IEEE, 2021.

\bibitem[Simon et~al.(2010)Simon, Meessen, and De~Vleeschouwer]{simon2010visual}
Cedric Simon, Jerome Meessen, and Christophe De~Vleeschouwer.
\newblock Visual event recognition using decision trees.
\newblock \emph{Multimedia Tools and Applications}, 50:\penalty0 95--121, 2010.

\bibitem[Peng et~al.(2023)Peng, Yin, Yang, Chen, and Lin]{multiviewclustering}
Siyuan Peng, Jingxing Yin, Zhijing Yang, Badong Chen, and Zhiping Lin.
\newblock Multiview clustering via hypergraph induced semi-supervised symmetric nonnegative matrix factorization.
\newblock \emph{IEEE Transactions on Circuits and Systems for Video Technology}, 33\penalty0 (10):\penalty0 5510--5524, 2023.
\newblock \doi{10.1109/TCSVT.2023.3258926}.

\bibitem[Huang et~al.(2023)Huang, Wang, and Lai]{fastmultiviewclustering}
Dong Huang, Chang-Dong Wang, and Jian-Huang Lai.
\newblock Fast multi-view clustering via ensembles: Towards scalability, superiority, and simplicity.
\newblock \emph{IEEE Transactions on Knowledge and Data Engineering}, 35\penalty0 (11):\penalty0 11388--11402, 2023.
\newblock \doi{10.1109/TKDE.2023.3236698}.

\bibitem[Arazo et~al.(2020)Arazo, Ortego, Albert, O’Connor, and McGuinness]{arazo2020pseudo}
Eric Arazo, Diego Ortego, Paul Albert, Noel~E O’Connor, and Kevin McGuinness.
\newblock Pseudo-labeling and confirmation bias in deep semi-supervised learning.
\newblock In \emph{2020 International Joint Conference on Neural Networks (IJCNN)}, pages 1--8. IEEE, 2020.

\bibitem[Chen et~al.(2023)Chen, Gao, Sun, and Sang]{chen2023ccsd}
Jiyuan Chen, Changxin Gao, Li~Sun, and Nong Sang.
\newblock Ccsd: cross-camera self-distillation for unsupervised person re-identification.
\newblock \emph{Visual Intelligence}, 1\penalty0 (1):\penalty0 27, 2023.

\bibitem[Buciluundefined et~al.(2006)Buciluundefined, Caruana, and Niculescu-Mizil]{originSelfDistillaion}
Cristian Buciluundefined, Rich Caruana, and Alexandru Niculescu-Mizil.
\newblock Model compression.
\newblock In \emph{Proceedings of the 12th ACM SIGKDD International Conference on Knowledge Discovery and Data Mining}, KDD '06, page 535–541, New York, NY, USA, 2006. Association for Computing Machinery.
\newblock ISBN 1595933395.

\bibitem[Redmon et~al.(2016)Redmon, Divvala, Girshick, and Farhadi]{redmon2016you}
Joseph Redmon, Santosh Divvala, Ross Girshick, and Ali Farhadi.
\newblock You only look once: Unified, real-time object detection.
\newblock In \emph{Proceedings of the IEEE conference on computer vision and pattern recognition}, pages 779--788, 2016.

\bibitem[Sibson(1973)]{SLINK}
R.~Sibson.
\newblock {SLINK: An optimally efficient algorithm for the single-link cluster method}.
\newblock \emph{The Computer Journal}, 16\penalty0 (1):\penalty0 30--34, 01 1973.
\newblock ISSN 0010-4620.

\bibitem[Reddy et~al.(2022)Reddy, Tamburo, and Narasimhan]{Reddy_2022_CVPR}
N.~Dinesh Reddy, Robert Tamburo, and Srinivasa~G. Narasimhan.
\newblock Walt: Watch and learn 2d amodal representation from time-lapse imagery.
\newblock In \emph{Proceedings of the IEEE/CVF Conference on Computer Vision and Pattern Recognition (CVPR)}, pages 9356--9366, June 2022.

\bibitem[Lin et~al.(2014)Lin, Maire, Belongie, Bourdev, Girshick, Hays, Perona, Ramanan, Doll{\'{a}}r, and Zitnick]{DBLP:journals/corr/LinMBHPRDZ14}
Tsung{-}Yi Lin, Michael Maire, Serge~J. Belongie, Lubomir~D. Bourdev, Ross~B. Girshick, James Hays, Pietro Perona, Deva Ramanan, Piotr Doll{\'{a}}r, and C.~Lawrence Zitnick.
\newblock Microsoft {COCO:} common objects in context.
\newblock \emph{CoRR}, abs/1405.0312, 2014.

\bibitem[Jocher et~al.(2023)Jocher, Chaurasia, and Qiu]{yolov8}
Glenn Jocher, Ayush Chaurasia, and Jing Qiu.
\newblock {Ultralytics YOLO}, January 2023.
\newblock URL \url{https://github.com/ultralytics/ultralytics}.

\bibitem[French(1999)]{catastrophicforgettingfrench1999}
Robert~M French.
\newblock Catastrophic forgetting in connectionist networks.
\newblock \emph{Trends in cognitive sciences}, 3\penalty0 (4):\penalty0 128--135, 1999.

\bibitem[Cheng et~al.(2022)Cheng, Wang, and Long]{meta-learningfewshot}
Meng Cheng, Hanli Wang, and Yu~Long.
\newblock Meta-learning-based incremental few-shot object detection.
\newblock \emph{IEEE Transactions on Circuits and Systems for Video Technology}, 32\penalty0 (4):\penalty0 2158--2169, 2022.
\newblock \doi{10.1109/TCSVT.2021.3088545}.

\bibitem[Wang et~al.(2023)Wang, Shi, Dong, Gao, Song, and Gong]{semantincknowledguideclassincrementallearning}
Shaokun Wang, Weiwei Shi, Songlin Dong, Xinyuan Gao, Xiang Song, and Yihong Gong.
\newblock Semantic knowledge guided class-incremental learning.
\newblock \emph{IEEE Transactions on Circuits and Systems for Video Technology}, 33\penalty0 (10):\penalty0 5921--5931, 2023.
\newblock \doi{10.1109/TCSVT.2023.3262739}.

\bibitem[Pillutla et~al.(2022)Pillutla, Kakade, and Harchaoui]{aggregationweights}
Krishna Pillutla, Sham~M. Kakade, and Zaid Harchaoui.
\newblock Robust aggregation for federated learning.
\newblock \emph{IEEE Transactions on Signal Processing}, 70:\penalty0 1142--1154, 2022.
\newblock \doi{10.1109/TSP.2022.3153135}.

\end{thebibliography}

\newpage
\end{document}